\documentclass[preprint,12pt]{elsarticle}

\usepackage{amssymb}

\journal{Information Fusion}

\begin{document}

\begin{frontmatter}


\author{Thomas Johnson}
    \ead{thomas.johnson@ntu.ac.uk}

 \author{Eiman Kanjo}
  \ead{eiman.kanjo@ntu.ac.uk}
 
  \author{Kieran Woodward}
    \ead{kieran.woodward@ntu.ac.uk}

\title{DigitalExposome: Quantifying the Urban Environment Influence on Wellbeing based on Real-Time Multi-Sensor Fusion and Deep Belief Network}




\address{}

\begin{abstract}
In this paper, we define the term 'DigitalExposome' as a conceptual framework that takes us closer towards understanding the relationship between environment, personal characteristics, behaviour and  wellbeing using multimodel mobile sensing technology.  Specifically, we simultaneously collected (for the first time) multi-sensor data including urban environmental factors (e.g. air pollution including: {\it PM1, PM2.5, PM10, Oxidised, Reduced, NH3 and Noise}, People Count in the vicinity), body reaction (physiological reactions including: {\it EDA, HR, HRV}, Body Temperature, {\it BVP} and movement) and individuals' perceived responses (e.g. self-reported valence) in urban settings. Our users followed a pre-specified urban path and collected the data using a comprehensive sensing edge devices. The data is instantly fused, time-stamped and geo-tagged at the point of collection.
A range of multivariate statistical analysis techniques have been applied including Principle Component Analysis, Regression and Spatial Visualisations to unravel the relationship between the variables. Results showed that {\it EDA} and Heart Rate Variability {\it HRV} are noticeably impacted by the level of Particulate Matter ({\it PM}) in the environment well with the environmental variables. Furthermore, we adopted Deep Belief Network to extract features from the multimodel data feed which outperformed Convolutional Neural Network and achieved up to ({\it a=80.8\% , {$\sigma$}=0.001}) accuracy.
\end{abstract}


\end{frontmatter}


\section{Introduction}
The long-term exposure to urban environment stressors such as particulate matter, gases and noise have been found to significantly impact an individual's behaviour and psychological health \cite{Kanjo2018}, \cite{Guite2006}, \cite{Bhat}. The World Health Organisation ({\it WHO}) found that 91\% of people are living in places where the air quality guidelines are not met and the use of non-clean fuels and household emissions in the atmosphere are causing over 4.2 million deaths each year \cite{Organisation2018}. In addition, those living in some locations in the UK have a higher risk of developing serious health conditions such as higher heart rate \cite{Kanjo2018}, asthma and cardio-cerebrovascular disease \cite{Stamatelopoulou2018} where a lifetime of exposure to high-levels of pollution can result in reduced life expectancy. \cite{Environment2019}. 

Recent developments in urban sensing and Internet of Things ({\it IoT}) has created the possibility to utilise environmental and on-body sensing tools to monitor the environment and its impact on individuals \cite{Kanjo2018}, \cite{Stamatelopoulou2018}.  Sensor-based technologies are becoming increasingly popular due to their availability to collect data in real-time, affordability and small size \cite{Larkin2017}, \cite{Debord2016}, \cite{Dai2017}, \cite{MaximilianUeberham2018}. These advances continue to enable more opportunities for capturing environmental signature in urban setting by providing the mechanisms to collect and analyse objective data \cite{Kanjo2018}, physiological changes \cite{Kanjoa} and behaviour markers of mental wellbeing \cite{Woodward}, \cite{MaximilianUeberham2018} in real-time. In addition, the major advances and recent developments within data science have created greater opportunities to understand large multimodel datasets through machine learning \cite{Kanjo2018}, deep learning \cite{Kanjoa} and spatial visualisations \cite{Johnson2020}.

In this paper, we define '{\it DigitalExposome}' as the quantification step in understanding the relationship between the environment and the body along with the perceived environmental responses which could potentially help in designing our cities with wellbeing in mind. 
In utilising the '{\it DigitalExposome}' concept, this paper leads us to explore, discuss and address the following research question: {\it How can we monitor, fuse, model and understand the person-environment interaction to help determine what makes an urban environment, potentially healthy}".
To answer this question we setup a data collection study in a busy urban area. Our participants collected (for  the  first  time) multi-sensor data simultaneously including:

\begin{enumerate}
\item Urban  environmental  attributes and air quality:  {\it PM1, PM2.5, PM10, Oxidised, Reduced, NH3 and Environmental Noise}
\item Body physiological reactions including: Heart rate (HR), Electrodermal activity ({\it EDA}), Heart Rate variability ({\it HRV}), Body temperature ({\it TEMP}) 
\item Body Movement via Accelerometer
\item People count via wireless proximity detection
\item Individuals’ perceived responses: Self-reported valence
\end {enumerate}

The data collection tools were built by the project team including a sophisticated sensing edge (Envro-Edge) with 10 embedded air quality sensors. A smart phone app (EnvBodySens2) that collects accelerometer data, Bluetooth Low Energy (BLE) signal for people count, self-report labels, Noise, Date/Time and GPS traces. On-body data was collected using E4 Empatica. The data is instantly fused, time-stamped and geo-tagged at the point of collection. By collecting the data in the "wild" and out of the lab paves the way for more realistic approach that can generalise to real-life setting.

To  the  best  of  our  knowledge  this combinatory and systematic data collection approach including a comprehensive list of on-body, contextual  and  environmental sensors along with the user responses has not attempted before. A range of multivariate statistical analysis techniques have been applied including Principle Component Analysis, Regression and spatial visualisations (including heat maps and geometrical tessellation) to explore correlated patterns in the data and unravel the association between the attributes which might suggest a causal relationship.  
Consequently, we identify common patterns arising from a group of people and mapping the patterns of sensor variance in a place with its stimuli. As a result of this, we can visualize the spatiality of wellbeing on three different levels: Individual wellbeing (the wellbeing of one individual in same environment - temporal), Accumulated wellbeing (the wellbeing of one individual in many environments - spatial), and Collective wellbeing (the wellbeing of group of individuals in many environments).

Furthermore, to delve into  the  predictive ability  of the heterogeneous multivariate attributes several machine  learning models were built and evaluated ranging from standard machine learning methods such as  K-Nearest  Neighbor,  Decision Trees and Support Vector Machines to more sophisticated deep neural networks-based techniques such as Convolutional Neural Network (CNN). We also adopted Deep Belief Network (DBN) to extract features from the multimodel data feed which then fed in to the machine learning algorithms. The performance of the on-body modality and environment modality is compared with and without DBN, assessing the possibility for a number of machine learning algorithms to infer affect quality (mental wellbeing) from the  data. 

\section{Related work}
Repeated and continuous human exposure to the environment and high-concentrated air pollutants have been found to increase the risk of developing serious conditions such as respiratory and cardiovascular diseases \cite{Lelieveld2020}, \cite{Lee2014} or even death \cite{TheGuardian}. Research recently has began focusing towards how the environment can impact physical health but it also is necessary to explore how the environment can impact mental wellbeing. 
Pollution within the urban environment is a continual problem contributing to rising health and mental wellbeing challenges. The ability to monitor air pollutants, physiology and mental wellbeing will enable the relationship between repeated environment exposures and mental wellbeing to be established.

ExpoApp \cite{Donaire-Gonzalez2019} used a sensor fusion approach (environmental and on-body) to model the short term health impact of high air pollution. Their analysis showed those who didn't have access to green spaces inhaled a higher rate of air pollution. A similar study monitored the environmental impact to an individual, indicating a positive correlation between the environment, body temperature, ElectroDermal Activity (EDA), motion and Heart Rate (HR) \cite{Donaire-Gonzalez2019}. In addition 'Project Helix' \cite{Maitre2018} studied the environmental impact on individuals living in urban environments. Increased levels of blood pressure, asthma, allergy related illnesses and behaviour issues were found for those living in urban environments. 

Mobile technology in previous research coupled with sensors have aimed to provide a deeper understanding into the impact of exposure to an individual in a particular location. This highlights the potential of recent technological advances, whereby an individual's exposure to the environment can be accurately assessed and calculated \cite{Stamatelopoulou2018}. Furthermore, particular areas have been found to have an increased risk of individuals developing serious health conditions such as higher heart rate \cite{Kanjo2018}, asthma and cardio-cerebrovascular disease \cite{Stamatelopoulou2018}. A study in 2018 used mobile technologies to develop the methods of assessing exposure to an individual. This involved using an activity and GPS sensor to predict an individual’s location. Overall the investigation demonstrated the capability of using sensors to accurately assess an individual’s exposure \cite{Stamatelopoulou2018}.

Personal sensors to measure individual exposure such as air pollution, noise, outdoor temperature, physical activity and blood pressure have been a positive way forward in monitoring due to their ability to collect data continually and in real-time \cite{Debord2016} helping to reveal early health conditions \cite{Nieuwenhuijsen2014}. By combining these sensor data streams together and the possibility for an individual to continuously wear sensors, the data can show the exposures an individual encounters as well as predict early health conditions.

\begin{figure}[ht]
\centering
  \includegraphics[width= 8cm, height= 5cm]{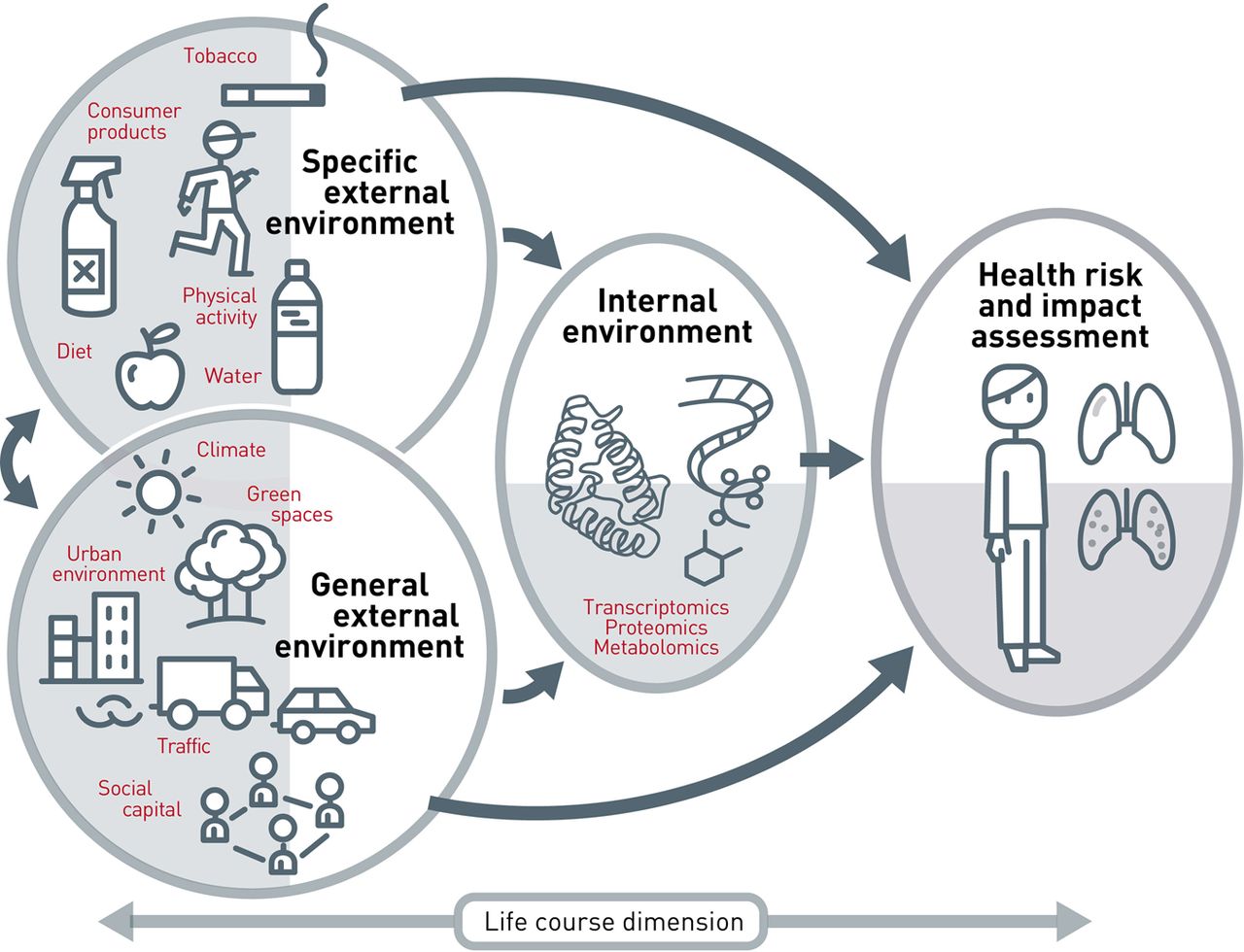}
  \caption{Demonstrating how the three stages of the exposome concept play a part in calculating the health risk \cite{Vrijheid2014}}
  \label{fig:exposome}
\end{figure}

Developed in 2005, the exposome concept encompasses each exposure that is subjected to a human from birth to death \cite{Wild}. In recent years, the concept is now actively being used in research communities as an alternative method to measuring the impact of the environment. High polluted environments have shown increased risk of developing conditions like asthma and cardio-cerebrovascular diseases \cite{MirandaLohDimosthenisSarigiannisAlbertoGottiSpyrosKarakitsiosAnjoekaPronk2017}. Figure \ref{fig:exposome} presents the exposome concept in its simplest stage and highlights the large amount of data that is required in order to calculate exposure impact across an individual lifetime. 

There are three stages associated with the exposome; internal, general exposome and specific external. The first stage of calculating the exposome is, ‘internal’ that measures the body’s biological response to exposures; such as ageing and stress. The second stage, ‘general exposome’ considers the wider impact on our lives and influences on the individual such as their education background and financial situation. Finally, the ‘specific external’ which examines effects out-side of the body such as air pollution, radiation and diet. Once all three stages have been measured, the exposome can be exactly calculated. 

In its current form researchers are able to calculate some of the impact of environmental exposure, however there still remains some challenges. Most studies have found it challenging to address and understand exposome fully because of its size, quantity of data \cite{SirouxValerie&AgierLydiane&Slama2016} and the overall quality of the data produced \cite{MirandaLohDimosthenisSarigiannisAlbertoGottiSpyrosKarakitsiosAnjoekaPronk2017}. 

The environment and its impact has been widely researched, however there has been little consideration of implementing a sensor-fusion approach using physiological and environmental sensors to study the impact of the environment on mental wellbeing.

\section{DigitalExposome}
\begin{figure}[ht]
\centering
  \includegraphics[width= 8cm, height=6cm]{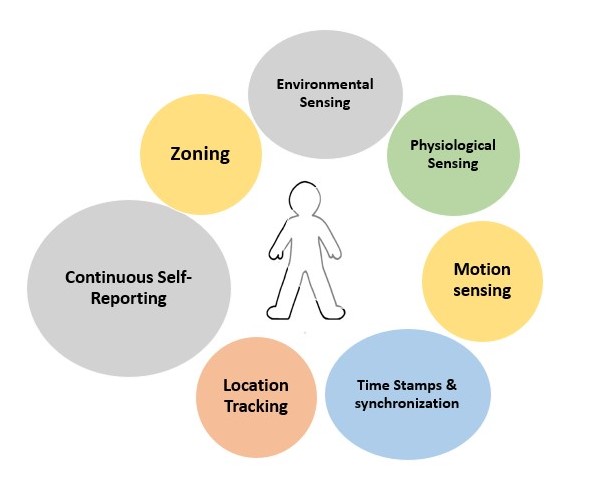}
  \caption{Data Collection methods to support {\it DigitalExposome}.}
  \label{fig:DExposomeConcept}
\end{figure}

We introduce the term '{\it DigitalExposome}' as a framework to quantify an individual's exposure to the environment by utilising a range of technological, mobile-sensing and digital devices, as shown in figure \ref{fig:DExposomeConcept}. This concept aims to measure multiple environmental factors using mobile technologies and then quantify them in real-life settings. Combining multiple data collection methods helps to support {\it DigitalExposome} and gain a better understanding into how exposures to the environment can impact mental wellbeing. 

This concept, further promotes to the use of the exposome concept by digitally providing a better understanding into the impact of exposure directly to an individual. Through \textit{DigitalExposome}, we aim to explore the opportunities that we for-see with this concept in exploring the link between pollution and wellbeing. To support this work, we have developed a range of sensing devices and applications. 

{\it DigitalExposome} is primarily made up of two parts: data collection and data analysis. Both aspects make use of technological advances in order to calculate the exposome. In order to quantify the process, we propose the utilisation of data from sensors that show how an individual has been exposed to pollutants. We see this as being a key part of the exposome concept, where both terms are clearly connected through their vision of being able to capture the true exposure that an individual has been exposed to. Data that is generated through the use of technology, such as sensors is ideal to monitor various exposures and enable the possibility to link this to health. 

\section{Methodology}
\subsection{System Architecture}
\begin{figure}[ht]
\centering
  \includegraphics[width= 10.5cm, height= 6cm]{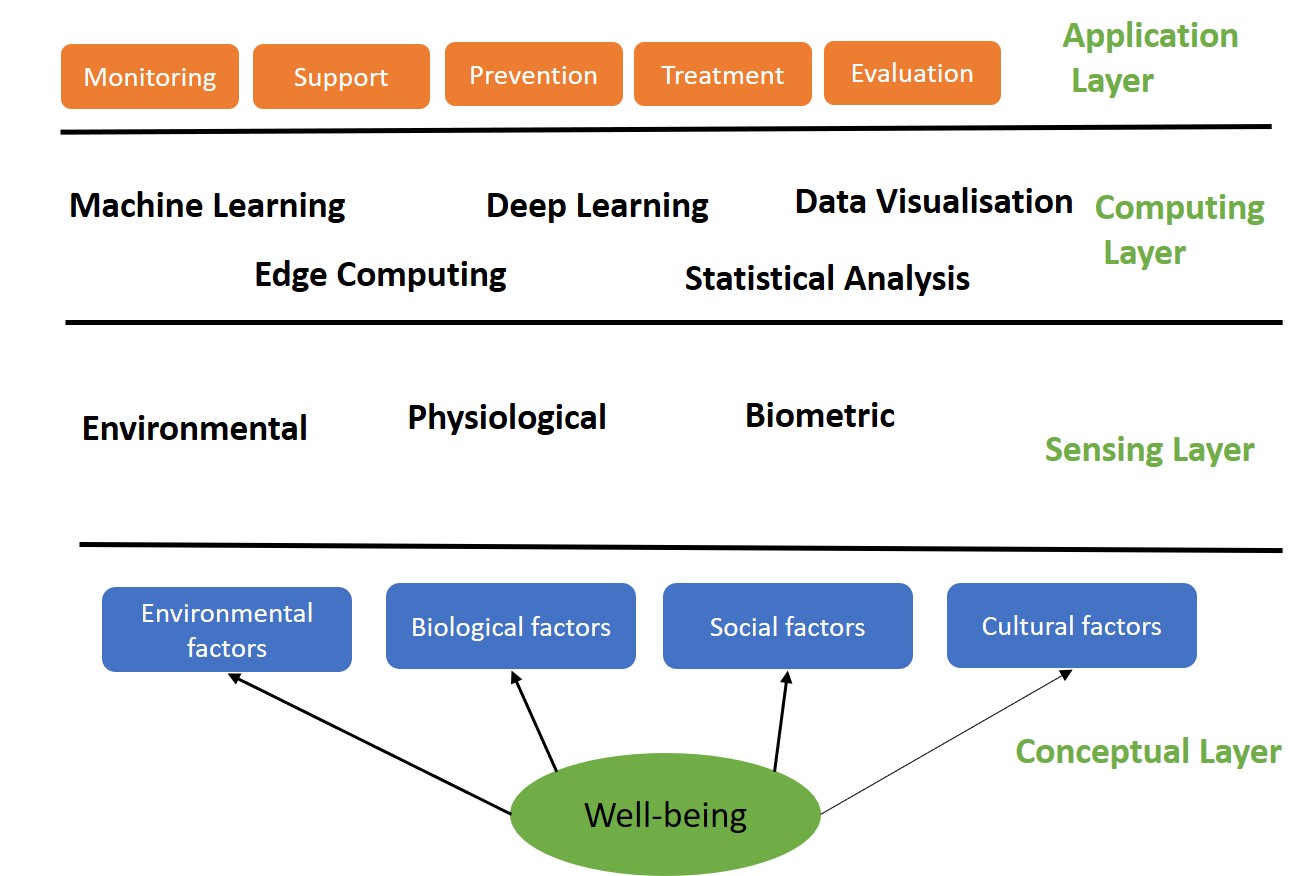}
  \caption{Conceptual and System Architecture of {\it DigitalExposome} for Wellbeing.}
  \label{fig:DExposomeSysArc}
\end{figure}

Figure \ref{fig:DExposomeSysArc} presents the conceptual system architecture of {\it DigitalExposome} with four key layers. Firstly, the conceptual layer explains the four main areas that can impact mental wellbeing include environmental, biological, social and cultural factors \cite{Liang2019}. The sensing layer contains the physical devices (e.g. smartphone and wristband) and physiological systems to monitor {\it HR, EDA} and body temperature along with the environmental factors such as air quality. The computing layer lists several key core data science techniques that enables processing and analysis of the data including: Machine Learning, Deep Learning, Statistical Analysis and Data Visualisation. Finally, the application layer presents potential applications of {\it DigitalExposome}. 

\subsection{Experimental Setup}
Following approval from Nottingham Trent University's Ethics Committee, we recruited a total of 20 participants (15 Males and 5 females) and carried out the study between September and October 2020. Due to the lockdown and COVID'19 restrictions it proved to be difficult to recruit further users. 

\begin{figure}[ht]
\centering
  \includegraphics[width= 13.5cm, height= 6.5cm]{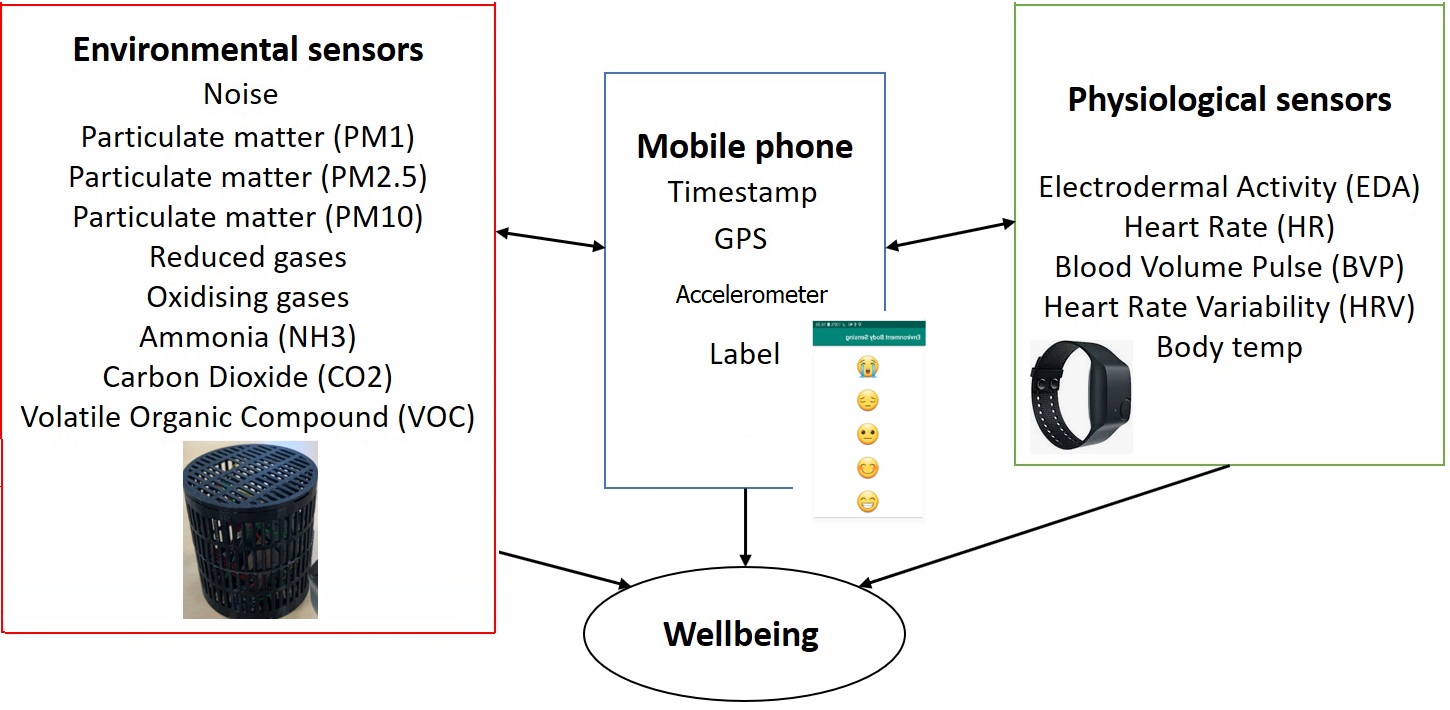}
  \caption{List of the fused variables collected by each device.}
  \label{fig:relationshipEnvirWell}
\end{figure}

Participants' were each provided with three devices; an environmental monitoring device (Enviro-Edge), Empatica E4 wristband and Samsung phone ready with the EnvoBodySends app. Each participant walked around a pre-specified route around Nottingham Trent University (Clifton Campus). Whilst walking, the three devices continually collected sensor data on environmental pollutants and physiological changes. In addition, participants self-reported their wellbeing continuously during the walk. The information acquired from each device is shown in Figure \ref{fig:relationshipEnvirWell}. 

The specified route took participants around 25 minutes to complete. The route was limited to 25 minutes due to user preference and not wanting to exhaust participants which could impact their body responses.

\begin{figure}[ht]
\centering
  \includegraphics[width= 7cm, height= 4cm]{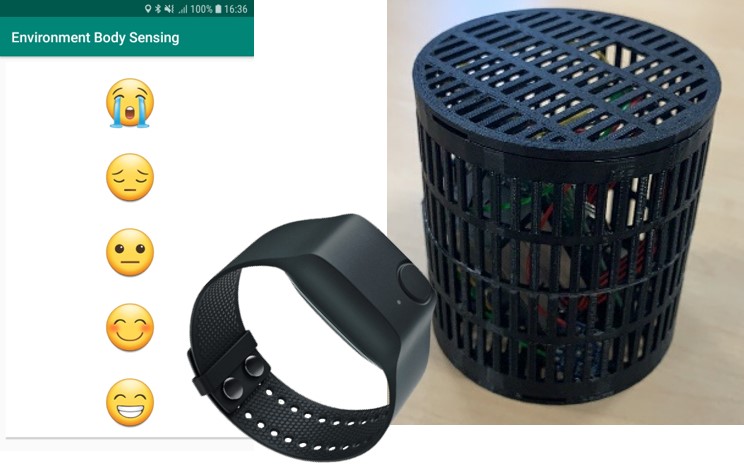}
  \caption{(left) Screenshot of smartphone application, (middle) {\it E4} Empatica, (right)Environment monitoring kit.}
  \label{fig:experimentsetup}
\end{figure}

The experiment data collection tools are depicted in Figure \ref{fig:experimentsetup} that include the Enviro-IoT, E4 Empatica and smartphone application. The Enviro-IoT edge device equipped with a Raspberry Pi 4 records environmental data continually once every 20 seconds. While the E4 Empatica sensors' data is sampled at different rates with {\it HR at 1Hz and EDA, BVP, HRV} and body temp at{\it 64Hz}. 

Each participant used the custom built pre-installed "EnvBodySens" smartphone app to record their perceived wellbeing. We have adopted the 'Personal Wellbeing Index for adults' which asks the user how they are feeling with their life as a whole \cite{Cummins}. This has been adapted in the form of a five-point Likert SAM scale \cite{Bradley1994} to provide a proven method for self-reporting subjective wellbeing. In our pre-installed mobile app the user is met with five well-know emojis from {\it 1=negative/low } to {\it 5=positive/high }, selected through the use of buttons throughout the walk. The idea is that the participant will be constantly prompted by the researcher to ascertain how they are feeling. Unfortunately, the application did not successfully log data for 8 of the participants resulting in only data from 12 participants being labelled. However, the unlabeled data can remain utilised for unsupervised statistical analysis.

\subsection{Pre-processing}
Following the data collection, the data was cleaned and pre-processed. Due to the varying sample rates, the physiological data collected ({\it EDA, BVP, HRV } and body temperature) were down-sampled to a rate of {\it 1Hz} to match the sample rate of collected {\it HR} by the device. In addition, the collected environmental sensor data had to be up-sampled to match the sampled rate of the physiological data at {\it 1Hz}. This was due to the low sample rate produced by the environmental device. Finally, the labelled data from the mobile smartphone was extracted and up-sampled to the same rate as the environmental and physiological data to  1Hz to remain consistent with the other data. To sample the data we have used linear interpolation \cite{Needham1959}. If the two known points are given by the coordinates ( x\textsubscript1 , y\textsubscript1), the linear interpolant is the straight line between these points. For a value x in the interval ( x\textsubscript2 , x\textsubscript1 ), the value y along the straight line is given from the equation of slopes 
as shown below: 

\begin{equation}
y=y_{1}+\left(x-x_{1}\right) \frac{\left(y_{2}-y_{1}\right)}{\left(x_{2}-x_{1}\right)}
\end{equation}

Following this, all signals were then normalised to bring all variables within the same range for both the data analysis and machine learning. Finally, all the sampled sensor data was fused together. 

Whilst cleaning the data, there were two variables excluded from the experiment; Carbon Dioxide and Volatile Organic Compound because of issues with logging the data resulting in no change in value during the experiment. In total the number of samples, after cleaning were 13,658.

\section{Results}
\subsection{Statistical Factorial Analysis}
\begin{figure}[ht]
\centering
  \includegraphics[width= \linewidth]{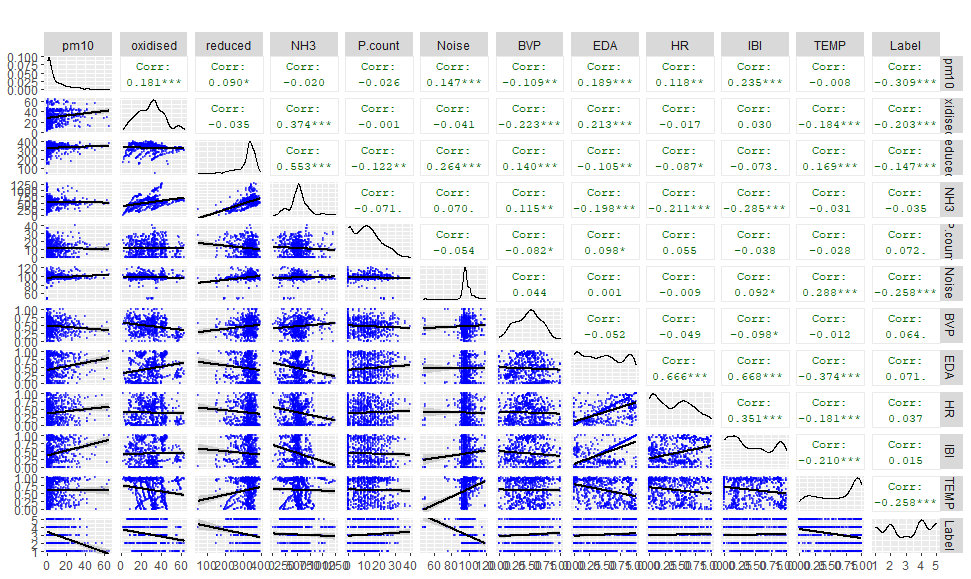}
  \caption{Correlation Matrix of the Environmental and Physiological Variables.}
  \label{fig:cmatrix}
\end{figure}

We have employed mathematical and statistical approaches for the exploratory analysis stage including variable Correlations, {\it PCA} factor maps, variable importance and Pearson’s R Correlation Coefficient to measure the association between two categorical  variables. A correlation matrix has been depicted at Figure \ref{fig:cmatrix} to further understand the relationship between the different variables. From the matrix, it is clear to see that some variables are highly correlated together. Analysing the individual cells shows {\it HRV} correlates well with PM10 and NH3. In addition, {\it EDA} demonstrates a correlation with PM10, Oxidised and Reduced gases and NH3. 

{\it PCA} Factor Maps are an effective method for large datasets, to help understand the relational impact between different variables, with reducing information loss \cite{Jollife2016}. Also, using {\it PCA} maps provides a visual method of presenting data and observing correlations between different variables \cite{Kanjo2018}. {\it PCA} factor maps give a view of all the variables projected on to a plane, spanned by the first two principle components. This method demonstrates the structural relationship between the different variables. We have demonstrated two {\it PCA} factor map plots based on variable importance of the many variables we have collected.

\begin{figure}[ht]
\includegraphics[width=14cm] {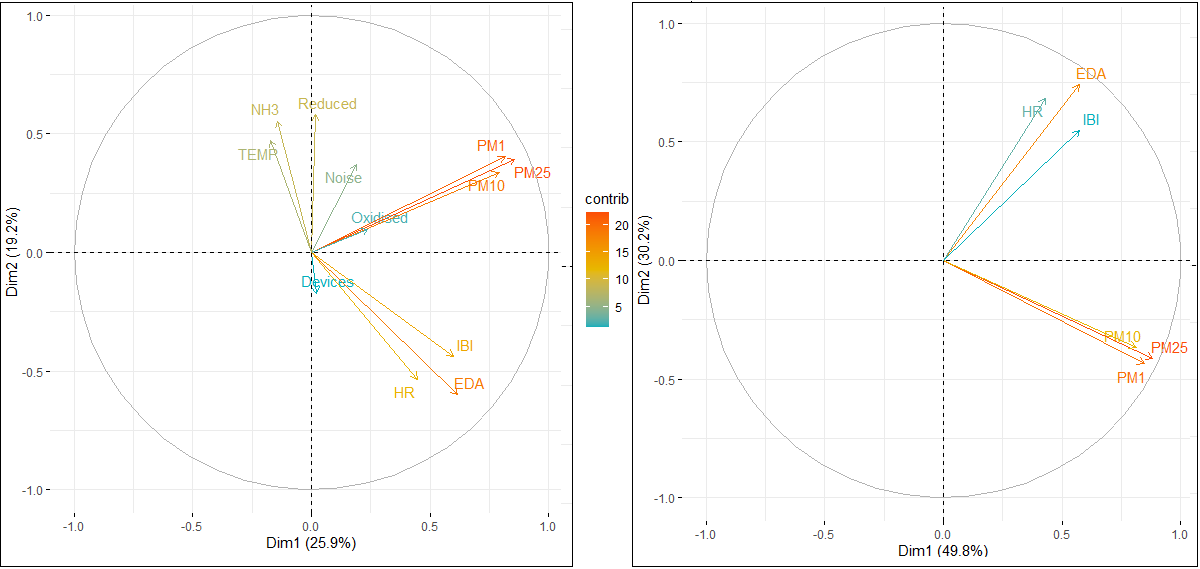}
  \caption{PCA Analysis - (left) Variance between the different variables, (right) Variance between the different variables without EDA.}
  \label{fig:pcamapall}
\end{figure}

Figure \ref{fig:pcamapall} (left), presents the captured environmental and physiological variables depicted on a {\it PCA} map. It is worth noting, that most of the body attributes {\it EDA, HR and HRV} are all at the top of the figure. Whilst, the environmental variables {\it PM1, PM2.5, PM10} and Reducing gases are located in the middle. From the diagram we can see that the first principle component explains about 25.9\% of the total variance and the second principle component an additional 19.2\%. Therefore, the first two principle components explain 45.1\% of total variance. The most important (or, contributing) variables are highlighted using the color gradient.

By removing the least important variables and keeping the most important (or relevant) ones which are positioned close to each others (including {\it PM1,PM10, PM2.5,EDA, HR and IBI}) we then notice that the two first components explain a large percentage of total variance, (first component explains around 50\% of the variance), as shown in Figure \ref{fig:pcamapall} (right). The close grouping and proximity of the independent variables suggests that {\it HRV, HR and PM10} are correlated and that {\it HRV, HR, PM2.5} are also being positively correlated.

\begin{figure}[ht]
\centering
  \includegraphics[width=8cm]{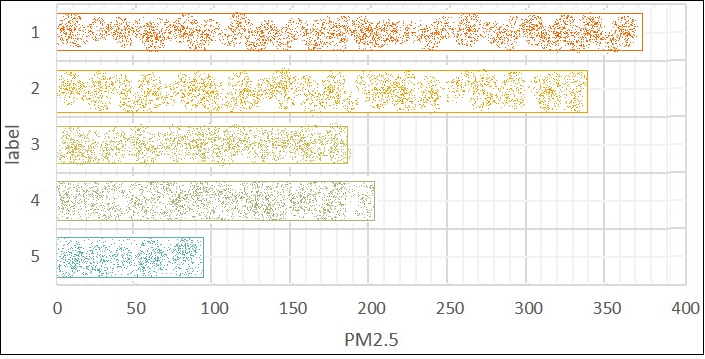}
  \caption{Depicts the relationship between the self-reported Participant's wellbeing (Label) and PM2.5.}
  \label{fig:pm25impact}
\end{figure}

Furthermore, Figure \ref{fig:pm25impact}, depicts an analysis on the environmental and physiological data indicated that where participants labelled their wellbeing as negative and very negative, there were high levels of PM2.5 within the environment. This is similarly true for PM1.0 and PM10 again demonstrating the direct correlation between human physiology and environmental pollutants.

\newpage
\subsection{Multi-Variant Regression Analysis}
Using PCA analysis and covariance matrix's allows us to explore the relationship unfolding between the different variables. We continue this process by using multi-variant regression to greater understand the importance and impact each variable has on the other. This work aims to study the variable dependency on two different modalities using Multivariate Regression and Principle Component Analysis (PCA). For each of the dependent variables (physiological data) we have used Multiple Linear Regression to compare against the independent variables (environmental data). The aim of this is to see which dependent variable can be predicted from using the environmental data as independent variables. 

\textbf{Multiple Regression Model for EDA:}
Firstly, we have used a multiple linear regression module for EDA to understand the impact of this physiological on-body sensor to the other independent environmental variables including NH3, Noise, PM1, PM2.5, PM10 and Reduced. Table \ref{MVR_EDA}, shows the multiple regression results for EDA: 

\begin{table}[h!]
\caption{Multiple Regression Analysis between EDA and Environmental variables.\label{MVR_EDA}}
\begin{tabular}{|l|r|r|r|r|}
\hline & Coefficients & Standard Error & t Stat & P-value \\
\hline Intercept & -0.02381894 & 0.02225209 & -1.07041 & 0.284452 \\
\hline nh3 & 0.000291595 & $1.14608 \mathrm{E}-05$ & 25.44285 & $1.5 \mathrm{E}-139$ \\
\hline noise & 0.004050864 & 0.000221511 & 18.2874 & $7.92 \mathrm{E}-74$ \\
\hline oxidised & -0.00590754 & 0.000143065 & -41.2928 & 0 \\
\hline pm1 & -0.00768185 & 0.00081832 & -9.38735 & $7.11 \mathrm{E}-21$ \\
\hline pm10 & 0.000939923 & 0.000285371 & 3.29369 & 0.000991 \\
\hline pm25 & 0.003698711 & 0.000800215 & 4.622149 & $3.83 \mathrm{E}-06$ \\
\hline reduced & -0.00058528 & $4.59985 \mathrm{E}-05$ & -12.7239 & $7.06 \mathrm{E}-37$ \\
\hline
\end{tabular}
\end{table}

\begin{figure}[ht]
\centering
  \includegraphics[width=9cm]{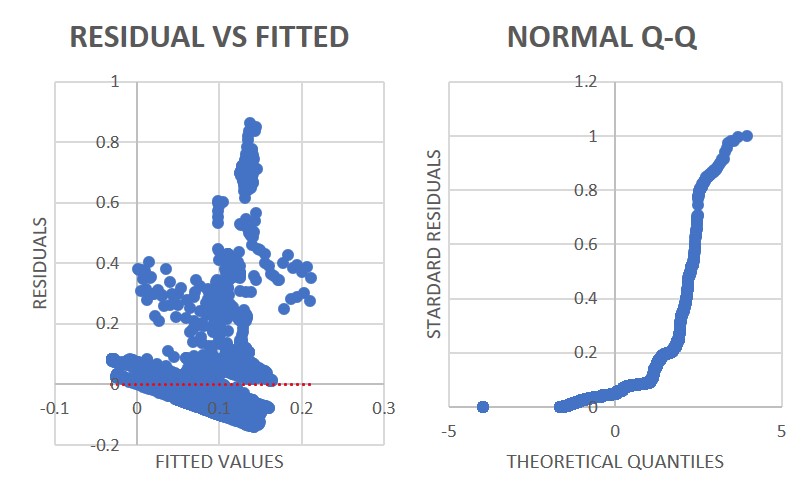}
  \caption{EDA regression curves.(Left) Residuals curve.(Right) Presents the Q-Q curve.}
  \label{fig:EDA_MVR}
\end{figure}

The data in Table \ref{MVR_EDA} was then evaluated using a regression curve shown in Figure \ref{fig:EDA_MVR}. This shows the relationship between the calculated residual values verses the fitted values. 

Right of Figure \ref{fig:EDA_MVR} shows the Normal Q-Q plot for EDA constructed by using bi-modal data. In many Q-Q plots, the data on the graph takes the shape of a twist like seen in this plot \cite{Kanjo2018}, \cite{DavidScott}. The lower part of the plot is almost linear, suggesting a normal distribution in relation to one mode of data distribution. In addition, the upper part of the Q-Q plot again suggests linear, showing an approximate distribution. The steep line between the upper and lower curve suggest that distribution is more dispersed than the distribution plotted.

\textbf{Multiple Regression Model for HR}

Below presents the multiple linear regression model for HR using the other independent variables (environmental). This includes; NH3, Noise, PM1, PM2.5, PM10 and Reduced. Table \ref{HR_EDA}, shows the multiple regression results for HR: 

\begin{table}[h!]
\caption{Multiple Regression Analysis between HR and Environmental variables.\label{HR_EDA}}
\begin{tabular}{|l|c|c|r|r|}
\hline & Coefficients & Standard Error & t Stat & P-value \\
\hline Intercept & 128.8420806 & 1.721454748 & 74.84488381 & 0 \\
\hline nh3 & 0.007322924 & 0.000886623 & 8.25933903 & $1.59864 \mathrm{E}-16$ \\
\hline noise & -0.083286817 & 0.017136432 & -4.860219147 & $1.1856 \mathrm{E}-06$ \\
\hline Oxidised & -0.051833849 & 0.011067698 & -4.683344891 & $2.84951 \mathrm{E}-06$ \\
\hline pm1 & 0.118538171 & 0.063306454 & 1.872450026 & 0.061165731 \\
\hline pm10 & 0.112184632 & 0.022076708 & 5.081583292 & $3.79248 \mathrm{E}-07$ \\
\hline pm25 & -0.232804742 & 0.061905795 & -3.760629225 & 0.000170194 \\
\hline reduced & -0.072042617 & 0.003558515 & -20.24513451 & $8.12597 \mathrm{E}-90$ \\
\hline
\end{tabular}
\end{table}

These initial findings are in agreement with previous research that shows PM2.5 can directly impact HR \cite{Paoin2020}. In addition, research has shown how differing levels of irregular environmental noise can impact a regular heart-beat. In particular, recent studies exploring this find that noise levels between 55 and 75 Decibels (dB) are linked to a higher risk of developing heart related diseases \cite{Munzel2014}.  

\begin{figure}[ht]
\centering
  \includegraphics[width=9cm]{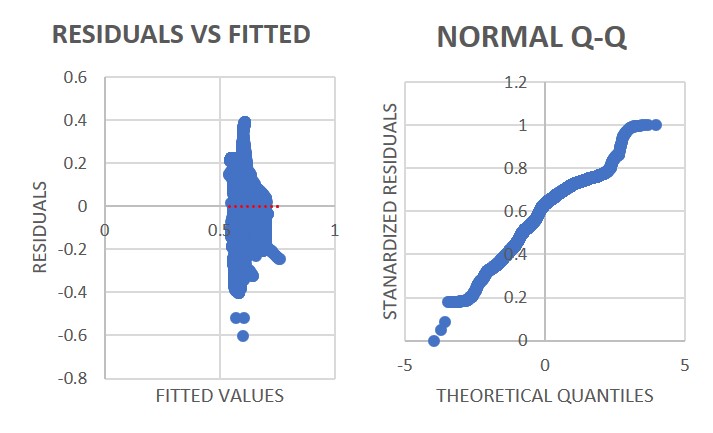}
  \caption{HR regression curves.(Left) Residuals curve.(Right) Presents the Q-Q curve.}
  \label{fig:HR_MVR}
\end{figure}

Similar to the EDA Q-Q plot, HR Q-Q plot shown in Figure \ref{fig:HR_MVR} (right) demonstrates a twist at either end of the plot. In addition the data shows a clear bi-modal distribution. The lower part of the plot is almost linear suggesting an approximate normal distribution. The line in the middle of the upper and lower parts follows a more linear (y=x) line, meaning that the distribution is less dispersed. It is worth noting that there were three outliers for HR distribution due to erroneous sensor readings.

\section{Visualisations}
\begin{figure}[ht]
\centering
  \includegraphics[width= 15cm, height=9cm]{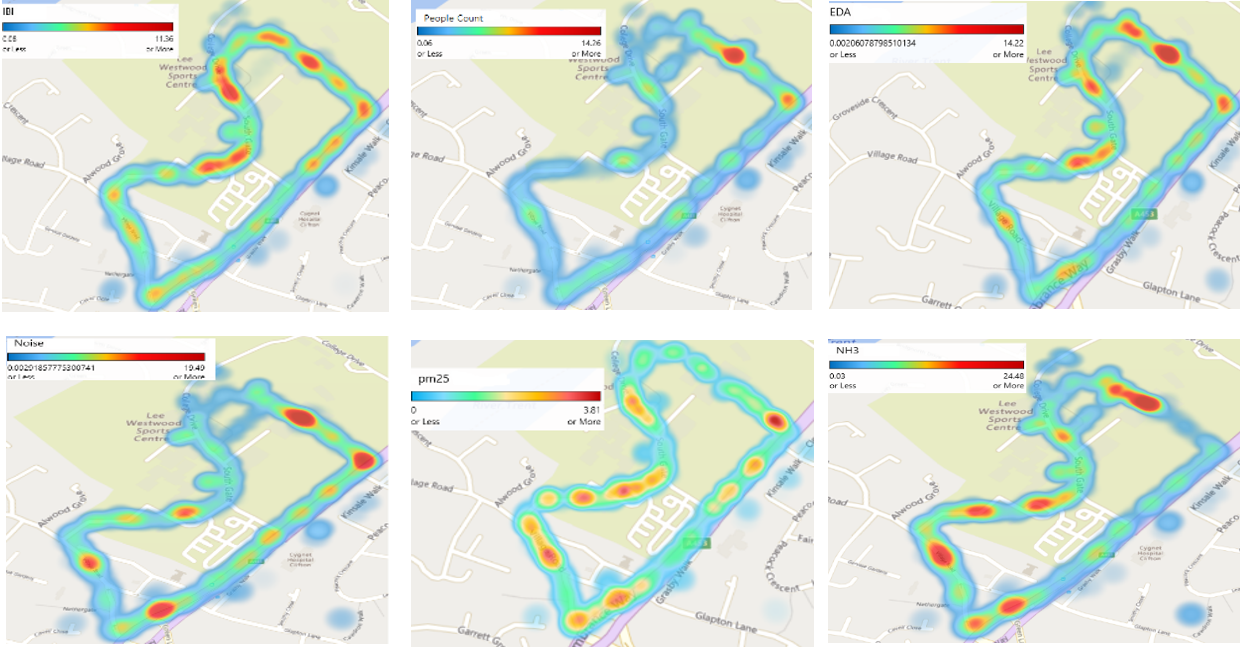}
  \caption{A heat map demonstrating environmental and physiological sensors data along the specified route.}
  \label{fig:heatmap}
\end{figure}

To summaries the dynamic sensing patterns and act upon the ﬁndings using visualisation, the geographical study area needs to be divided in smaller areas. One common way of looking at patterns is to use heat maps of the sensor data. For example, Figure \ref{fig:heatmap} presents several heat maps plotting environmental and physiological sensor data, showing the changes whilst the participant travelling along the route. In particular, (upper right of the map) we can observe that moving from the campus towards the main road, each participant is subjected to an increase of PM2.5 and Noise and was met with an increase of HRV and EDA. This approach further demonstrates the impact of the environment on mental wellbeing states. 

It can be seen that the sensor data hot-spots are scattered along the path. While the heat maps show the level of intensity based on GPS traces coordinates, the sensor data on these heat maps indicate the real distribution of sensor data. One option is to divide the study area into grid cells \cite{Kanjo2010} however, it is difficult to allocate a cell to each sensor reading, moreover, it is not possible to decide on the cell size, since the density of the sensor mobility traces can be of different density distribution.

To address these issues, we utilise combinatorial computational geometry algorithm called “Voronoi”, which is a diagram partitioning of a plane into regions based on distance to points in a specific subset of the plane \cite{Dobrin}. The method of Voronoi visualisations is a computational geometry algorithm which allows the visualisation of large data sets \cite{Dobrin}. The concept works by defining a set of polygon regions called cells, whereby the cells give an indication of the overall density of an object area of the size of the object itself \cite{Pokojski2018}.

\begin{figure}[ht]
\centering
  \includegraphics[width= 10cm,height=7cm]{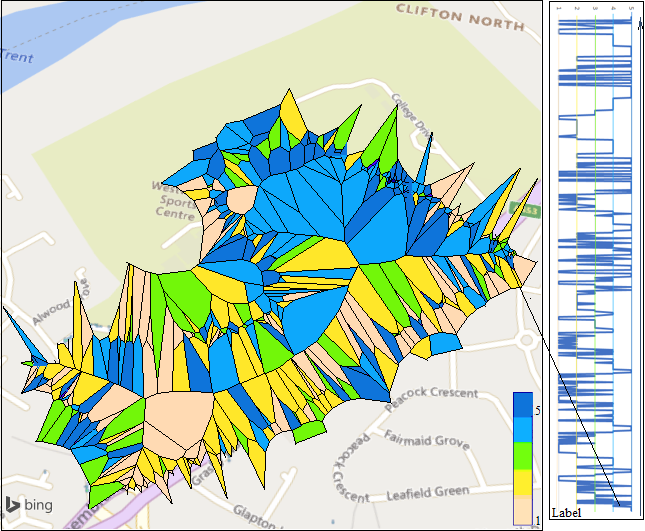}
  \caption{(left) Voronoi overlay from one participant data. Each polygon represents one location trace tagged with a wellbeing label while collecting the data in specified route (the map layer from Microsoft Bing), (right collected label data from  start to end).}
  \label{fig:Voronoi}
\end{figure}

Voronoi Diagram divides the space into a set of regions called Voronoi cells, including the space that is closest to the object (route location, in our case). The size of these cells gives an indication of the density of the area a certain object is in or the size of an object\cite{Pokojski2018}. The cell structure also shows the Delaunay triangulation, which easily allows calculating an object's immediate set of neighbours.
The definition of a Voronoi cell is given by the following equation, where {\it x} is a planar metric space; {\it p} is the set of generator points in the metric space; and d is the distance between all points in x and a specific generator point (where the distance can be defined using any distance definition such as Euclidean, Manhattan, or road-network distance):
\begin{equation}
Vor{_i}=\left \{ x \mid d(x,p{_i})\leq d(x,p{_j}),j\neq i \}\right.
\end{equation}

Thus, the Voronoi diagram is composed of a collection of tessellations (i.e. polygons) defined as Vor, where:
\begin{equation}
Vor{_i}=\left \{ Vor{_1}, Vor{_2}... Vor{_n}\}\right.
\end{equation}

The creation of a Voronoi tessellations is a dynamic procedure till all the points are represented in adjacent polygons. If sufficient number of particles did not satisfy Equation (1) then Voronoi gets partially filled. In this case, the data is then redistributed. By giving each polygon a class value Ci that corresponds to the sensor value collected in a particular GPS coordinate, it is then possible to divide the space into adjacent polygons with different sensor reading which are represented in colours.

Figure \ref{fig:Voronoi}, presents the self-reported wellbeing data using the app on the specified route for this experiment. The color of the polygons represents the wellbeing data from low negative to high positive. The visualisation demonstrates that poor wellbeing (lighter colour) was most reported along the main road where high levels of pollution were also experienced whereas more positive states of wellbeing was recorded in less polluted areas such as fields and open spaces.

\section{Deep Learning Classification}
The use of machine learning and deep learning networks have been explored to classify the five self-reported states of wellbeing using the pollution and physiological data from the 12 participants who successfully labelled their wellbeing. 

\subsection{Deep Belief Network}
Deep learning presents many opportunities to classify raw sensor data using classification models such as Convolutional Neural Networks ({\it CNNs}) but deep learning is mostly effective for deep feature extraction. To enable depth level feature extraction from the fused environmental and physiological data we employ an unsupervised one dimensional deep belief network ({\it DBN}  \cite{Hinton2006}, \cite{Fischer2014}, \cite{Hassan2019}.

Unsupervised {\it DBNs} are beneficial as they learn to extract a deep hierarchical representation of the training data which can then be used as features within a supervised machine learning classifier. {\it DBNs} are generative models and are a composition of stacked Restricted Boltzmann Machines ({\it RBM}) and Sigmoid Belief Networks \cite{Mocanu2018}, \cite{Smolander2019}. {\it RBMs} are stacked and trained in a greedy manner by training in a sequential way, feeding lower layers’ results to the upper layers to form {\it DBNs}. They model the joint distribution between the observed vector x and the $\ell$ hidden layers $h^{k}$ where $x=h^{0}, P\left(h^{k-1} \mid h^{k}\right)$ is the distribution of the units conditioned on the hidden units of the {\it RBM} at level $k$, and $P\left(h^{\ell-1}, h^{\ell}\right)$ is the visible-hidden joint distribution in the top {\it RBM}:

\begin{equation}
P\left(x, h^{1}, \ldots, h^{\ell}\right)=\left(\prod_{k=0}^{\ell-2} P\left(h^{k} \mid h^{k+1}\right)\right) P\left(h^{\ell-1}, h^{\ell}\right)
\end{equation}

Unsupervised learning is used to train the {\it RBMs} of the {\it DBN} to automatically construct features and reconstruct inputs. The Gibbs Sampling based contrastive divergence method used to train the {\it RBM} is shown below:

\begin{enumerate}
\item The fused physiological and pollution sensor data is fed into the RBM as the input $x=h^{(0)}$ of the first layer.
\item  The activation probabilities of the hidden layers are calculated using (2):
\begin{equation}
P\left(h_{j} \mid X\right)=\sigma\left(b_{j}+\sum_{i=1}^{m} W_{i j} X_{i}\right)
\end{equation}
\item The activation probabilities of input layers are calculated using (3):
\begin{equation}
P\left(X_{i} \mid h\right)=\sigma\left(a_{i}+\sum_{j=1}^{n} W_{i j} h_{j}\right)
\end{equation}
\item The edge weights are updated where $\alpha$ is the learning rate using (4):
\begin{equation}
W i j=W i j+\alpha\left(P\left(h_{j} \mid X\right)-P\left(X_{i} \mid h\right)\right)
\end{equation}
\end{enumerate}

After training the first {\it RBM} the edge weights are frozen and the remaining {\it RBMs} are trained using the same contrastive divergence method with the output of previous trained {\it RBM} being used as the input of the next {\it RBM}. After training has completed, the {\it DBN} features are extracted from the top hidden layer and a hidden unit of the learned network structure is used as the input layer for a supervised {\it ML} models. The {\it DBN} is essentially used as a feature selection mechanism for the machine learning models as it is used as a representation learner compressing the original input vector for the {\it ML} models to use.

\subsection{Results}
The extracted features from the {\it DBN} were combined with Random Forest, Support Vector Machine ({\it SVM}), Decision Tree, Gaussian Naive Bayes, Logistic Regression and Gradient Boosted supervised machine learning models to classify the five self-reported states of wellbeing using the pollution ({\it PM1, PM2.5, PM10, Oxidised, Reduced, NH3 and Noise}) and physiological ({\it BVP, EDA, HR, HRV} and body temperature) data. These machine learning models were selected due to their high popularity and were also tested using only common statistical features: mean, median, max, min, max-min, standard deviation and quartiles \cite{Lisetti2004a}. Additionally, a Convolutional Neural Network ({\it CNN}) has been trained using the same raw data to enable comparison with the {\it DBN} models. The models were trained over 20 epochs with a batch size of 128 and tested using 10-fold cross validation.

\begin{figure}[ht]
\centering
  \includegraphics[width= 12cm]{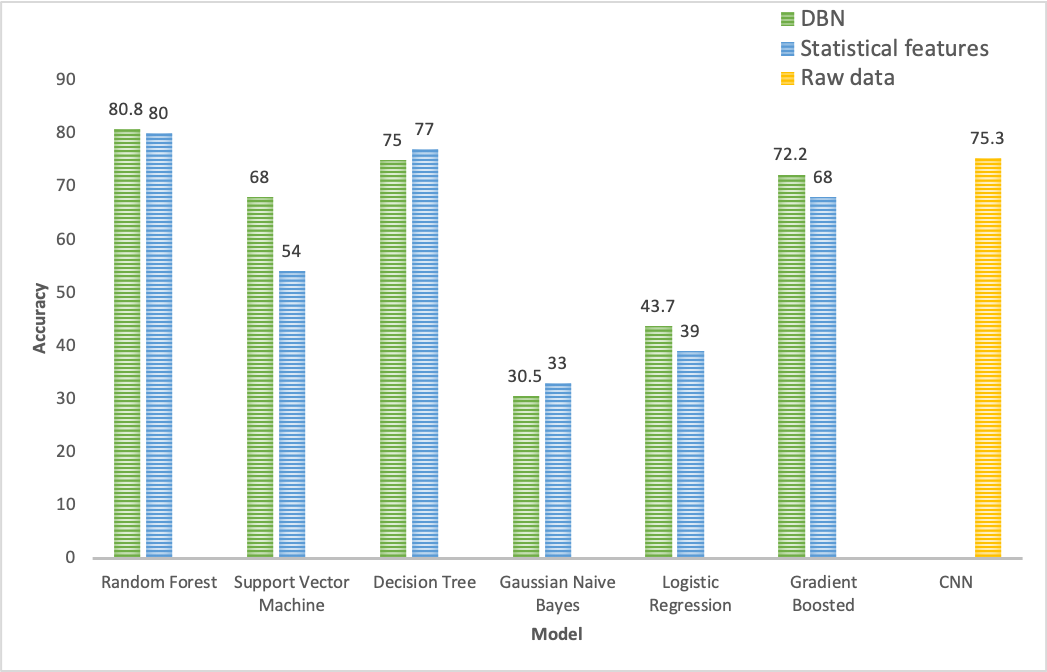}
  \caption{Comparison of classification models trained using statistical features and features extracted from DBN.}
  \label{fig:comparison}
\end{figure}

Figure \ref{fig:comparison} shows the accuracy for each of the classification models trained using standard statistical features and using features extracted using the {\it DBN}. The results demonstrate that the models trained using features extracted from the {\it DBN} outperformed the models trained with statistical features for three out of the five classifiers and achieved on average 3.2\% higher accuracy. Random Forest combined with the {\it DBN} was the best performing model achieving 80.83\% accuracy, outperforming all statistical models and the {\it CNN} which is frequently used for wellbeing classification by 5.54\%.

\begin{figure}[ht]
\centering
  \includegraphics[width= 8cm]{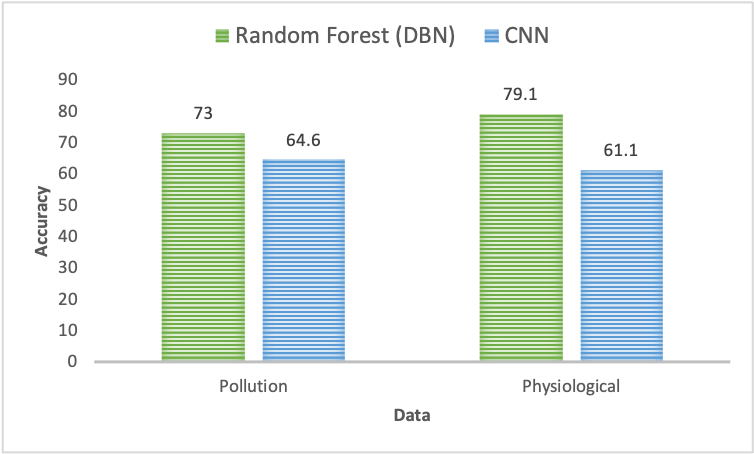}
  \caption{Comparison of Random Forest combined with DBN and CNN when trained using only pollution or physiological data}
  \label{fig:comparison1}
\end{figure}

To further explore the impact pollution has on wellbeing the best performing model (Random Forest) in addition to the {\it CNN} were individually trained using only the pollution and physiological data as shown in Figure \ref{fig:comparison1}.

The results show that wellbeing can be inferred using pollution data alone with 73\% accuracy while wellbeing can be inferred from physiological data with 79.1\% accuracy. It was expected that psychological would accurately classify wellbeing due to its high correlation with the sympathetic nervous system \cite{Sharma2012}. However, it is surprising that pollution data combined with physiological data outperformed the model trained using pollution data alone. Furthermore, the CNN trained using only pollution data outperformed the CNN trained using physiological data, suggesting pollutants have a considerable impact on wellbeing.

\section{Discussion and Limitations}
The 'DigitalExposome' concept can help to observe changes in the environment and its impact on human body at the same time. To the best of our knowledge, this is the first real-world study that has been shown to quantify the link between the environment and the impact on physiology and mental wellbeing.

Continually collecting and fusing real-world environmental and physiological sensor data helped us learn about our surroundings, how we interact and behave in different environmental conditions. This has gone beyond previous work in this area which typically only observes how noise can impact wellbeing \cite{Kanjo2010} and does not consider other environmental pollutants. 

A PCA analysis suggests that when all collected variables are combined together they can describe the variability of the data as a whole. In particular, on the PCA map, the physiological sensors ({\it EDA, HR and HRV}) point towards a different location to the environmental variables. From our analysis we can conclude that a range of environmental factors {\it PM1.0, PM2.5, PM10} impact physiological changes {\it HRV, HR}. 

Voronoi visualisations have given an indication of how changes within the environment can have an impact on mental wellbeing. Typically, it was found that where air pollution such as {\it PM1}, 2.5, 10 and Noise was increasing, participants labelled their wellbeing as very negative. This demonstrates consistent results with previous studies in this area \cite{Kanjo2010}, \cite{Johnson2020}. This form of spatial analysis, greatly helps in understanding the degree to which a place is similar to other nearby places. 

The ability to classify the collected data presents many possibilities for the real-world inference of wellbeing using pollution data. The results show that {\it DBNs} successfully improve the accuracy in which wellbeing can be inferred, compared with CNNs and machine learning models trained using statistical features. Combining physiological with pollution data achieved up to 80.8\% accuracy compared with 79.1\% when trained using only physiological and 73\% when trained using only pollution data. The ability for pollution data to increase overall accuracy demonstrates its impact on wellbeing and shows pollution should continue to be considered as a factor that influences changes in wellbeing. 

During this study some limitations were encountered. Early analysis on the collected sensor data found that the Empatica E4 was not reliably collecting participants' {\it EDA}. While the EDA sensor worked successfully for some, for other participants no variation in {\it EDA} was recorded throughout the experiment. At the point of fusing the collected sensor data, both CO2 and VOC were found to have collected data for some participants but not all resulting in its dismissal.
As this study was conducted during the {\it COVID-19} pandemic, collecting data in a participant heavy experiment was challenging task. In the future, we aim to recruit more participants to further investigate and generalise the relational impact of the environment on mental wellbeing.

\section{Conclusion and Future Work}
Although very appealing, the exposome approach is complex, both in terms of observing each factor, quantifying it and analysing its relationship to health.
This novel research challenges existing approaches to understanding the 'Exposome' concept, providing a new perspective on how to quantify the approach between the environment and mental wellbeing. The 'Exposome' concept was first proposed to encompass the totality of human environmental exposures from conception onward. This concept has then drawn the attention to the need for more multivariate data enabled by the rise of mobile and sensor technologies. 
This enables us to study how people experience place, the impact of place and the role that different environmental factors play on people.
In this paper, we proposed the new concept '{\it DigitalExposome}' that demonstrated the potential of employing a multi-model mobile sensing approach to further understand the relationship between the environment and it's impact on mental wellbeing. To achieve this, a real-world experiment was conducted around Nottingham Trent University, Clifton Campus where participants walked around a specified route reporting their responses and collecting environmental, behavioural and on-body sensor data.  

Statistical analysis including PCA, Multi variant Linear Regression, Voronoi and data spatial visualisations were implemented to explore the variation in data and the factor importance. We found that physiological (on-body) sensor data is directly correlated to pollution (PM in particular) within the environment. In addition, {\it DBNs} have helped successfully classify five states of wellbeing with up to 80.8\% accuracy using the fused physiological and pollution data. 

In the future, we hope to consider additional environmental sensors to observe greater changes that may improve our sense of places and  characterize the relationship between people and spatial settings, which in turns might influence the future design of urban spaces. Although the distance walked by participants was sufficient, greater distances offer a deeper observation on how longer periods of time within an urban environment has on changes in mental wellbeing states. In addition, we want to take our approach and apply it to a busier city centre environment to further understand the impact of heightened pollution on wellbeing.  
It is important to understand when and where they feel better, so they appreciate their surroundings and city planners create places where people feel rejuvenated. Sensing technology can shed the light on  how people breath, feel and interact with their environment in different surroundings. This can help in offering a better security for city dwellers and creating a bond with their environments.



  \bibliographystyle{elsarticle-num} 
  \bibliography{sample-base}





\end{document}